\newif\ifArxiv
\Arxivtrue
\newif\ifPlos
\Plosfalse

\documentclass[10pt,letterpaper]{article}
\ifPlos\usepackage[top=0.85in,left=2.75in,footskip=0.75in]{geometry}\fi
\ifArxiv\usepackage[top=0.85in,left=2.75in,footskip=0.75in]{geometry}\fi

\usepackage{amsmath,amssymb}

\usepackage{changepage}

\usepackage{textcomp,marvosym}

\usepackage{cite}

\usepackage{nameref,hyperref}

\usepackage[right]{lineno}

\usepackage[nopatch=eqnum]{microtype}
\DisableLigatures[f]{encoding = *, family = * }

\usepackage[table]{xcolor}

\usepackage{makecell}%because of the table
\usepackage{array}
\newcolumntype{L}[1]{>{\raggedright\arraybackslash}p{#1}}
\usepackage{booktabs}

\newcolumntype{+}{!{\vrule width 2pt}}

\newlength\savedwidth

\newcommand\thickhline{\noalign{\global\savedwidth\arrayrulewidth\global\arrayrulewidth 2pt}%
\hline
\noalign{\global\arrayrulewidth\savedwidth}}

\ifPlos\raggedright\fi
\usepackage[aboveskip=1pt,labelfont=bf,labelsep=period,justification=raggedright,singlelinecheck=off]{caption}

\makeatletter
\renewcommand{\@biblabel}[1]{\quad#1.}
\makeatother

\usepackage{lastpage,fancyhdr,graphicx}
\usepackage{epstopdf}
\fancyheadoffset[L]{2.25in}
\fancyfootoffset[L]{2.25in}
\ifArxiv\fi

\begin{document}
\vspace*{0.2in}

\begin{flushleft}
{\Large
\textbf\newline{Humans can learn to detect AI-generated texts, or at least learn when they can't} %or at least learn their own limits % Please use "sentence case" for title and headings (capitalize only the first word in a title (or heading), the first word in a subtitle (or subheading), and any proper nouns). 
}\newline
\\
Jiří Milička\textsuperscript{1},
Anna Marklová\textsuperscript{1*},
Ondřej Drobil\textsuperscript{1\Yinyang},
Eva Pospíšilová\textsuperscript{2\Yinyang}
\\
\bigskip
\textbf{1} Department of Linguistics, Faculty of Arts, Charles University, Prague, Czech Republic
\\
\textbf{2} Institute of Czech Language and Communication Theory, Faculty of Arts, Charles University, Prague, Czech Republic
\\
\bigskip

% Insert additional author notes using the symbols described below. Insert symbol callouts after author names as necessary.
% 
% Remove or comment out the author notes below if they aren't used.
%
% Primary Equal Contribution Note
\Yinyang These authors contributed equally to this work.

% Additional Equal Contribution Note
% Also use this double-dagger symbol for special authorship notes, such as senior authorship.
%\ddag These authors also contributed equally to this work.

%\textcurrency Current Address: Dept/Program/Center, Institution Name, City, State, Country % change symbol to "\textcurrency a" if more than one current address note
% \textcurrency b Insert second current address 
% \textcurrency c Insert third current address


% Group/Consortium Author Note
%\textpilcrow Membership list can be found in the Acknowledgments section.

% Use the asterisk to denote corresponding authorship and provide email address in note below.
* anna.marklova@ff.cuni.cz

\end{flushleft}
\section*{Abstract}
This study investigates whether individuals can learn to accurately discriminate between human-written and AI-produced texts when provided with immediate feedback, and if they can use this feedback to recalibrate their self-perceived competence. We also explore the specific criteria individuals rely upon when making these decisions, focusing on textual style and perceived readability.

We used GPT-4o to generate several hundred texts across various genres and text types comparable to Koditex, a multi-register corpus of human-written texts. We then presented randomized text pairs to 254 Czech native speakers who identified which text was human-written and which was AI-generated. Participants were randomly assigned to two conditions: one receiving immediate feedback after each trial, the other receiving no feedback until experiment completion. We recorded accuracy in identification, confidence levels, response times, and judgments about text readability along with demographic data and participants' engagement with AI technologies prior to the experiment.

Participants receiving immediate feedback showed significant improvement in accuracy and confidence calibration. Participants initially held incorrect assumptions about AI-generated text features, including expectations about stylistic rigidity and readability. Notably, without feedback, participants made the most errors precisely when feeling most confident --- an issue largely resolved among the feedback group.

The ability to differentiate between human and AI-generated texts can be effectively learned through targeted training with explicit feedback, which helps correct misconceptions about AI stylistic features and readability, as well as potential other variables that were not explored, while facilitating more accurate self-assessment. This finding might be particularly important in educational contexts, since the ability to identify AI-generated content is highly desirable and, more importantly, false confidence in this domain can be harmful.

\ifPlos\linenumbers\fi
\section{Introduction} 
In most everyday situations, individuals typically have some idea about their abilities, even if these ideas are imprecise and correlate only loosely with actual competence \cite{kruger1999unskilled,jansen2021rational}. For example, people who have never learned to swim are usually aware of this inability and understand that swimming can be learned. This awareness arises from experiences involving informal or formal testing, or from social interactions where they observe the difficulty of acquiring a skill. Such self-assessment capacities, however, tend to fail when confronted with entirely novel phenomena, such as interacting with artificial intelligence (AI). This is primarily because there are no standardized assessments, institutional oversight, or established instructional methods for these emerging skills, leaving intuition without any reliable anchor.

We may reasonably assume that most people have never undertaken a rigorous test to assess their ability to discriminate between human-written texts and texts generated by frontier language models. In the best case, individuals are agnostic about their skill levels; in the worst case, they hold unrealistic expectations that remain uncorrected. Consequently, many individuals neither perceive the need to learn this skill nor, even if they did, would they find clear guidance in the literature about whether or how such learning could be successful. False confidence in this domain can be particularly problematic in educational contexts --- for example, when a teacher is overly confident in identifying AI-generated texts and unjustly accuses students of academic dishonesty.

Our current study contributes to the extensive body of literature examining the conditions under which people can discern AI-generated texts from human-written ones. However, our central research question shifts focus from mere discriminatory ability to the capacity of individuals to utilize feedback effectively: Can individuals learn to accurately discriminate between human-written and AI-produced texts when provided with immediate feedback? Furthermore, are they capable of using this feedback to recalibrate their self-perceived competence?

Beyond these core questions, our research also explores the specific criteria individuals rely upon when distinguishing between human and AI-generated texts, identifying factors that aid or hinder accurate judgments. We also aim to identify the factors that are influenced by the learning process due to the feedback loop. We specifically focus on textual style (register), hypothesizing that people attribute particular stylistic characteristics to AI-generated texts. These assumptions might help if aligned with reality but could also lead to errors if based solely on prejudice. For similar reasons, we investigate perceived readability, analyzing how participants associate readability with text authorship. Additionally, we collected data on demographic variables (age, gender, education) and participants' engagement with AI technologies, including usage frequency and general attitudes.

For our experimental design, GPT-4o generated a corpus comprising 672 texts across various genres and text types, chosen specifically to reflect stylistic diversity comparable to the Koditex corpus, a traditional human-authored collection designed explicitly for genre and register richness \cite{zasina2018koditex}.

These texts were presented pairwise to participants (each receiving 17 randomly selected text pairs plus 3 control items). We recorded not only their accuracy in identifying which text was AI-generated versus human-written but also their confidence levels, response times, and judgments about text readability.

Participants (n=254) were randomly assigned to two experimental conditions. One group received immediate feedback after each trial, indicating correctness, whereas the second group received no immediate feedback and learned their overall results only upon completing the experiment.

All texts and participants were Czech native speakers, reflecting our intention to study a medium-sized language, as English disproportionately dominates the training datasets of contemporary language models, making it unrepresentative of other languages. Nevertheless, all data, experimental software, analytical scripts, and detailed protocols are publicly available, ensuring that the study can be easily replicated in other linguistic contexts.

\subsection{State of the art}
The question of whether humans can detect AI-generated texts has attracted significant scientific attention, but experimental research with human participants has shown mixed results. As the studies are incomparable in almost every aspect, we can only guess whether the differences in findings can perharps be attributed to the purpose of the study, selected text genres, differing populations tested, mode of presentation, language, or AI models used. 

To date, studies have reported varying overall accuracy in participants' recognition of AI-generated texts. \cite{cheinetal_2024} found that English-speaking participants performed slightly better than chance and attested substantial variation based on individual abilities or expertise. In other studies\cite{clarketal_2021, fleckensteinetal_2024,franketal_2024}, however, human evaluators’ accuracy was rather at chance levels.

Studies exploring how succesfully can people detect or distinguish AI-generated texts typically focus on only one genre, which corresponds to the general framing of the study (e.g., concern for a given field, such as cheating in education), or comparisons of distinct genres. Research has been done on poetry \cite{kobis_mossink_2021}, EFL students' essays \cite{fleckensteinetal_2024, dewilde_2024}, scientific abstracts \cite{gaoetal_2023}, news articles, recipes and short stories \cite{clarketal_2021}, social media posts \cite{radivojevicetal_2024}, or job applications, online dating and AirBnB host profiles\cite{jakeschetal_2023}. 

However, no previous studies have systematically addressed text variability as a continuum across registers, despite some research indicating that certain linguistic features related to text variability may influence how accurately people distinguish between AI- and human-generated texts. \cite{cheinetal_2024} applied the Linguistic Inquiry and Word Count 2022 toolbox to measure word-level characteristics of the text but did not find any significant impact. \cite{jakeschetal_2023} used both computational methods and humans to annotate texts based on certain language features, finding that while some of them are predictive of AI-content, some are not. However, humans still falsely rely on them and may therefore misjudge the given texts as AI-generated. We tried to address the impact of text variability on the ability to distinguish AI and human-generated texts, using a multi-dimensional register analysis framework.

To our knowledge, there has been no study targetting humans' ability to identify AI texts on Czech or other similarly `small' language. Most of the mentioned studies were conducted on English, except \cite{franketal_2024}, which tested humans' ability to recognize AI-generated content across texts, audio, and images. The authors compared the performance of speakers of English, German and Chinese, evaluating texts in their native tongue. They found that German speakers were less likely to evaluate AI-generated text as written by humans than US participants. They attribute this difference to the limited training data available for German texts. In case of languages like Czech, that has even more limited training data pool, we might expect this effect to be even more pronounced.

Studies varied in how texts were presented to participants. In some cases, authors employed a paired presentation mode in which one human-written and one AI-generated text were presented side-by-side and the evaluators were required to distinguish between the two \cite{kobis_mossink_2021}. Other studies presented single texts without explicit pairs, asking evaluators to categorize each text independently as human-written or AI-generated \cite{clarketal_2021, fleckensteinetal_2024, radivojevicetal_2024}. \cite{cheinetal_2024} combined both approaches and found that when texts are presented as a pair, the overall accuracy is higher. In our study, we will present texts in side-by-side mode.

As for the source of texts, the reviewed studies relied on large language models from OpenAI and employed different methods of prompting and creating AI-generated content. \cite{clarketal_2021} compared the performance based on different models of GPT, finding that the ability to detect AI-generated texts decreased between GPT-2 and (newer) GPT-3. As for prompting, for example,\cite{gaoetal_2023} created scientific abstracts by prompting ChatGPT with titles and the designated journal, whereas \cite{kobis_mossink_2021}  generated poetry using initial lines of human-written poems. \cite{fleckensteinetal_2024} provided the model with standardised criteria of what the EFL texts should look like. \cite{dewilde_2024} prompted the models with the same instruction as the EFL students who wrote the reference texts, but additionally manipulated the instructions to adjust the level of proficiency in English. 

As our study considers individual differences in participants, related work by \cite{cheinetal_2024} provides some relevant insights, finding that fluid nonverbal intelligence significantly predicts detection accuracy, whereas executive functioning, empathy or frequency of using smartphones and being online do not. \cite{franketal_2024} found that the degree to which people generally trust other people or institutions, cognitive reflection and familiarity with deepfakes also significantly affected accuracy.

\cite{fleckensteinetal_2024} compared performance between novice and experienced teachers evaluating EFL students' essays. Both groups were rather overconfident about their assessment. Novice teachers were generally unable to distinguish between AI- and human-generated texts, regardless of their quality. Experienced teachers performed slightly better when evaluating high-quality texts, but they were unable to correctly classify texts of lower quality. The authors attributed these differences to the fact that while experienced teachers may use their advanced knowledge of what texts should look like and are aware of patterns produced by AI, they fail to distinguish low-quality text because they do not realise that AI may downplay their performance and produce such low-quality texts.

As far as we know, there are no studies so far that investigate whether humans are capable of learning to discriminate between AI and human-written texts through feedback. The closest to our study is \cite{clarketal_2021}, who examined whether this task could be improved through brief instruction before the experiment. They either provided participants with a) instructions about what cues are relevant and which are rather misleading, b) examples of AI- and human-generated content with explanation of relevant cues, or c) a pair of correctly labeled texts that could be compared, again with explanation of relevant cues. They found that while all methods improved subsequent performance, only training with examples had a significant effect.

The main difference between our study and theirs is that in our experiment, learning can occur throughout the entire experiment, and most importantly, we offer no explanations nor recommend any specific cues. Instead, we allow participants to find their own cues, whether at a conscious or unconscious level. 

In the following section, we describe in greater detail the register variation in Czech.

\subsection{Register variation}
In the present study, we focus on linguistic variability that functionally contributes to the text composition. This variability has been a center of attention in the methodology developed by Douglas Biber \cite{biber1988variation}, which aims to interpret the variability according to several dimensions of variation, which then point out clusters of texts that are similar in those characteristics. Such clusters of texts are called registers. A register can be defined as “recurring variation in language use depending on the function of language and on the social situation” \cite{pescuma2023situating}.

In Czech, the register variation has been thoroughly examined by Cvrček et al. \cite{cvrcek2020registry}, using the methodology of multidimensional analysis (MDA). This methodology was first introduced by Biber \cite{biber1988variation} and it was adapted to the specific of the Czech language. From the analysis, 121 features were projected onto 8 dimensions of variation of Czech. These dimensions were interpreted in accordance with the features and text types that accumulate on their poles (\ref{tab:dimensionsCzech}).

\begin{table*}[htbp]
  \centering
  \caption{\bf Dimensions of variation in Czech.}
  \label{tab:dimensionsCzech}
  \begin{tabular}{|c| L{0.35\linewidth}| L{0.35\linewidth}|}
    \hline
    {\bf Dim.} & {\bf Positive pole} & {\bf Negative pole} \\
    \thickhline
    1 & dynamic & static \\\hline
    2 & spontaneous & prepared \\\hline
    3 & higher level of cohesion & lower level of cohesion \\\hline
    4 & polythematic & monothematic \\\hline
    5 & higher amount of addressee coding & lower level of addressee coding \\\hline
    6 & general/intension & particular/extension \\\hline
    7 & prospective & retrospective \\\hline
    8 & attitudinal & factual \\\hline
  \end{tabular}
\end{table*}

The Czech MDA examined linguistic variation using the Koditex corpus \cite{zasina2018koditex}. Koditex is a 9-million-word synchronic corpus of Czech, developed for exploring register variability. It encompasses diverse communication modes --- written, spoken, and internet-based --- each subdivided into specific divisions and classes, such as blogs, general fiction, or elicited speech. Rich annotations, such as lemmatization, morphological tagging, and named entity recognition, made it suitable for deep corpus analysis. Koditex comprises text samples of comparable length rather than full texts, making it well suited for MDA. The full results of the Czech MDA are available online at \url{https://jupyter.korpus.cz/shiny/lukes/mda/} \cite{lukes_mdavis}.

In this study, we used shortened original Koditex texts and their AI generated counterparts (more in \ref{sec:methodology}). We conducted a new MDA on these texts.

\section{Methodology}
\label{sec:methodology}
\subsection{Main research questions and hypotheses}
\label{subsec:researchquestion}
The objective of this study is to address the following research questions:

\begin{enumerate}
    \item Are native speakers of the Czech language capable of distinguishing AI-generated texts from texts that are human-generated?
    \item Does immediate feedback enhance the ability of Czech speakers to differentiate between AI-generated and human-generated texts?
    \item Does immediate feedback help the Czech speakers to update their confidence level so that it reflects their abilities to differentiate between AI-generated and human-generated texts?    
    \item Is the ability to determine whether a text is written by AI or a human influenced by the genre of the text?
    \item Is the ability to determine whether a text is written by AI or a human influenced by the percieved readability of the text?    
    \item Can the ability to determine text authorship be influenced by an individual's attitudes toward AI?
    \item Can the ability to determine text authorship be influenced by the frequency with which an individual uses AI in their daily life?
\end{enumerate}

Our hypotheses are as follows: 
\begin{enumerate}
    \item Czech speakers, when presented with two texts, one written by a human and one AI generated, will be able to guess which is which with higher than chance accuracy.
    \item Individuals who receive immediate feedback after each trial in the aforementioned task will demonstrate higher accuracy than those who do not receive feedback.
    \item Individuals who receive immediate feedback after each trial in the aforementioned task will demonstrate stronger positive relationship between confidence level and accuracy than those who do not receive feedback.
    \item The accuracy of guessing whether a text was written by a human or AI will be influenced by the stylometric qualities of the texts. The question of which qualities influence the accuracy is exploratory.
    \item  Human-written and AI-generated texts will differ in percieved readability measured during the experiment. The question of how percieved readability influence the authorship assessment correctness is exploratory.
    \item Individuals with a more positive attitude toward AI will be more successful in determining authorship.
    \item Individuals who interact with AI more frequently will be more successful in determining authorship.
\end{enumerate}

\subsection{Material}

For the preparation of language material, we used the Koditex corpus \cite{zasina2018koditex}. Due to its broad stylistic diversity, Koditex enabled us to explore the ability to determine authorship across different genres. Each text in Koditex was divided into two parts, with the first 500 words paired with the system prompt (in English): ”Please continue the Czech text in the same language, manner and style, ensuring it contains at least five thousand words. The text does not need to be factually correct, but please make sure it fits stylistically.” The texts were then generated using the GPT-4o-2024-05-13 model with a temperature of 0 (2024/6/30), trimmed to begin and end with complete sentences while maintaining approximately 100 words, and cleaned of various formatting characters with standardized quotation marks. The second part of the original human-written text underwent identical trimming and cleaning.

\begin{figure}[!htp]
    \centering
    \ifArxiv\includegraphics[width=\linewidth]{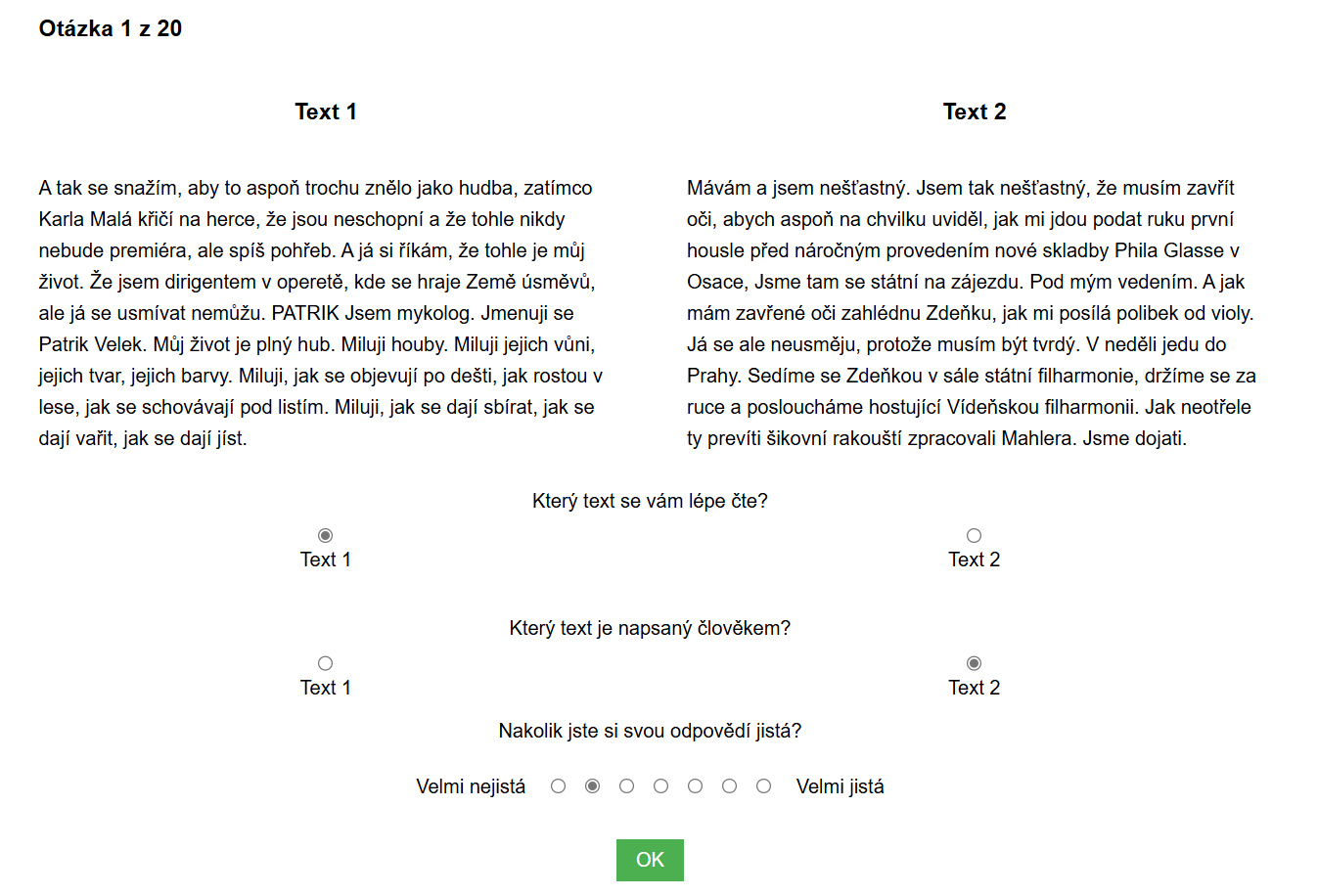}\fi
    \caption{{\bf Example of an experimental trial.} The questions under the two texts can be translated as: Which text reads more easily? Which text is written by a human? How confident are you in your answer?}
    \label{fig:example_trial}
\end{figure}

This process created 672 pairs of topically and stylistically comparable text chunks. From these pairs, one was randomly selected for each experimental trial (see \ref{fig:example_trial} for an example).

In addition, we manually selected three pairs of text chunks and created a control question attached to them, which was displayed on a separate screen after the trial. This question focused on the topic of the respective texts and was intended to determine whether participants remained attentive throughout the experiment. The question asked what topic the previous pair of texts addressed, and participants could choose from four options. For example, one of the control pairs of texts clearly described the Gobi Desert, and the options for the topics were `education', `Czech history', `desert', and `libraries'. These attention checks were excluded from the analysis and were used only to filter out inattentive participants from the dataset.

\subsection{Experimental design}

The study was conducted online through a dedicated application. Respondents received a link to the experiment, allowing them to complete it in the comfort of their home or another quiet environment. The experiment, including instructions and the demographic survey, was in Czech. Participants were instructed in the initial guidelines to complete the experiment on a computer, rather than a smartphone or tablet, which helped to ensure, as much as possible, comparable laboratory conditions. 

Firstly, participants were presented with a text on the screen displaying the title of the experiment, its basic description, and the contact information of the researchers conducting the study. After clicking the `Continue' button, participants were shown a link to the informed consent document in a PDF format. They were instructed to read it and then indicate their consent to participate by checking the box ``I have read the Informed Consent and I agree" before proceeding further.

This was followed by a demographic questionnaire, in which participants provided the following information: gender, age, mother tongue(s), presence of reading and text comprehension disorders, and information about their level of education, along with a field of study. %TOHLE BYCH MOŽNÁ I VYNECHALAIn the study section, participants could choose from the following levels of education: primary school, secondary school (with or without a high school diploma), higher vocational school, bachelor's degree, master's degree, doctoral studies, and other. In the field of study, they could select one or more of the following options: Philological (Czech studies, linguistics, translation, language teaching, literary studies...); Other humanities and social sciences (history, psychology, sociology, economics, arts, law, philosophy...); Computer science, machine learning; Other natural and technical sciences (biology, chemistry, physics, healthcare fields...); Other/Unable to categorize. Certain fields were categorized separately, as we assumed that studying these fields would lead to enhanced text analysis skills or advanced skills in working with artificial intelligence (for example philological fields, Czech language studies, computer science, and machine learning). 

After filling out these basic demographic details, participants indicated on an 8-point scale how often they work with artificial intelligence (ranging from every day to never) and completed a battery of 9 questions assessing their attitudes toward AI. The attitudes were tested in a form of statements; participants were asked to indicate on a 7-point Likert scale how much they agree with the following nine statements:
\begin{enumerate}
    \item Artificial intelligence can improve the quality of our everyday lives.
    \item Artificial intelligence poses a risk to human safety.
    \item Automatically generated texts can be as high-quality as texts written by humans.
    \item Automatically generated texts lack human creativity and personal style.
    \item Artificial intelligence is not capable of true understanding in the way that humans are.
    \item Artificial intelligence produces grammatically correct texts.
    \item I prefer to avoid using artificial intelligence when writing texts that matter to me.
    \item I am concerned that people are becoming dumber due to the use of artificial intelligence.
    \item I am fascinated by what artificial intelligence can accomplish when writing texts.
\end{enumerate}

After reading the brief instructions, participants were presented with a practice trial on the screen, featuring comments describing the course of the experimental task. Participants were randomly assigned to one of two groups. The first group received feedback after each trial, enabling them to learn from their mistakes (feedback group), while the second group received feedback only at the end of the experiment (no feedback group). 

\begin{figure}[!htp]
    \centering
    \ifArxiv\includegraphics[width=\linewidth]{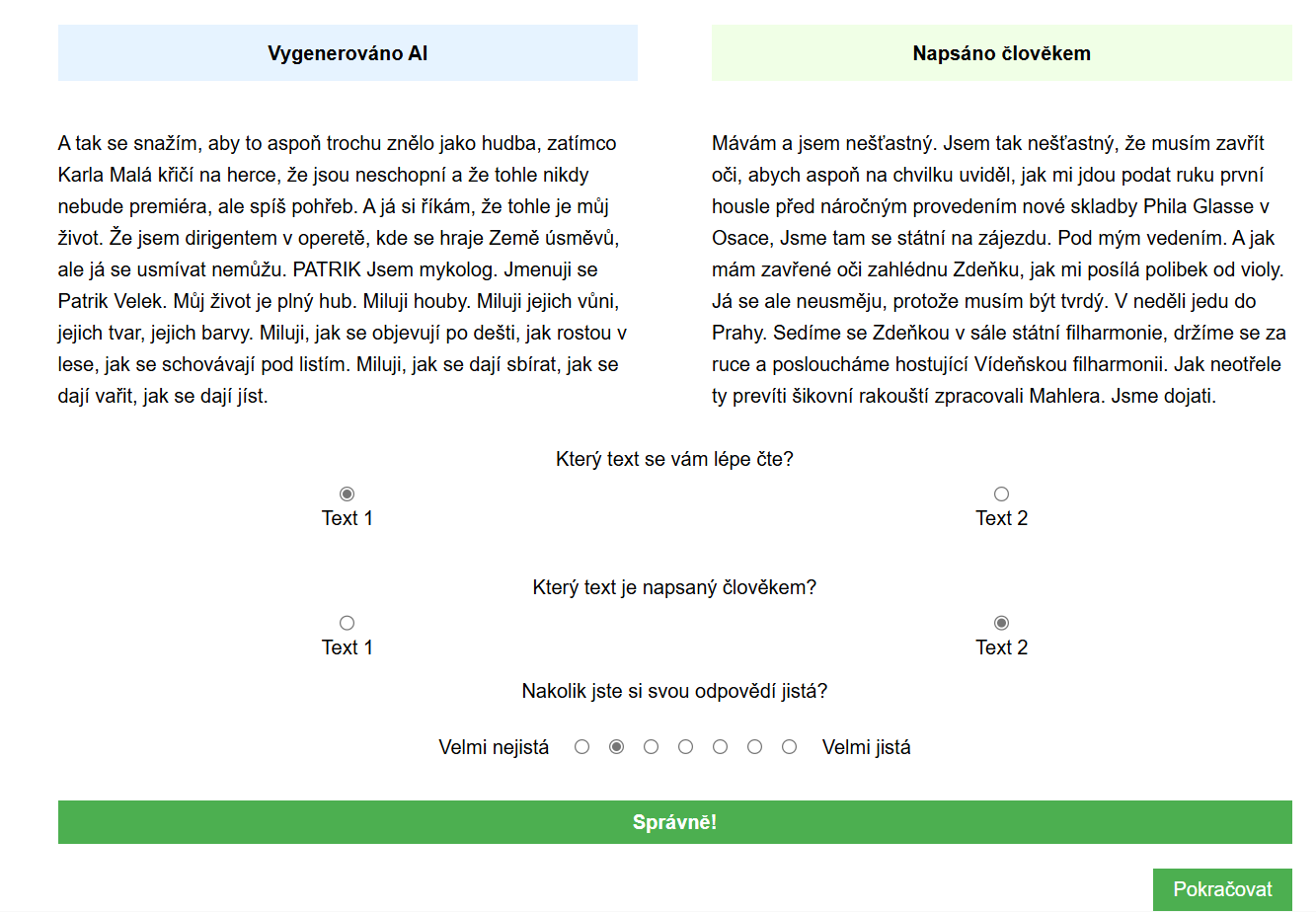}\fi
    \caption{{\bf Example of feedback after a correct answer.} The texts at the top say `Generated by AI' and `Written by human', and the text in the green window at the bottom says `Correct!'}
    \label{fig:example_correct answer}
\end{figure}

In each trial, participants were first presented with two texts in two columns, with one text written by a human and the other by artificial intelligence (see the layout of the experimental trial in Fig \ref{fig:example_trial}.
Along with the texts, participants were shown a question asking which of the two texts they found easier to read. After answering, participants were presented with a question asking which text was written by a human, as well as a 7-point Likert scale on which they had to indicate how confident they were in their authorship identification. If the participant was in the feedback group, after pressing the OK button in the trial, they were shown their success rate on the screen, accompanied by a clear identification of which text was written by a human and which by artificial intelligence (for an example of feedback for a correct answer, see Fig \ref{fig:example_correct answer}).

Participants completed a total of 20 trials. Of these, 17 were texts randomly selected from the corpus, and three were attention-check texts (positioned at trials number 3, 13, and 18) designed to verify participants’ attentiveness.  

At the end of the experiment, both participant groups (the experimental group, with feedback after each trial, and the control group) were shown a summary of their individual results. 

\subsection{Participants}
A pilot study was conducted prior to the actual experimental testing during the Open House Day at the Faculty of Arts, Charles University, Prague, with additional participants recruited later. The authors approached high school students, their parents, and other Open House visitors, inviting them to participate in the experiment. After each session, one of the authors conducted a brief interview with the participant to gather feedback on their experience and the technical aspects of the experiment. A total of 24 people participated in the pilot study. Following the pilot, several minor details were adjusted, but the overall experimental design remained unchanged. Data from the pilot study were not used in the main analysis.

The main sample was primarily recruited from the participant pool managed by the LABELS psycholinguistic laboratory. The pool consists of volunteers interested in participating in linguistic experiments and students who were offered university course credits for their participation. The dataset also includes several participants from the general public who responded to Facebook posts by the experimenters, accounting for approximately 22 participants. The recruitment started 25.02.2025 and ended 14.03.2025. The study was approved by the etical committee of the Charles University beforehand. 

The complete sample comprised 291 participants. However, 33 participants were excluded for failing to answer all attention-check questions correctly (the experiment was demanding, and some students participating for course credits showed insufficient attention). An additional 3 participants were excluded for not being native Czech speakers and one for low age (under 18 years old). 

The final sample consisted of 254 participants (female: 180, male: 70, other or preferred not to say: 4), the mean age was 24.11, with participants ranging from 19 to 80 years old. The majority of participants had a bachelor's (99) or master's (135) degree as their highest level of education, 14 participants were doctoral graduates, and only a few had completed only highschool (3), vocational (2) or other (1) type of education. The majority of participants had an educational background in social sciences and humanities (107), substantially represented were also students and graduates of natural sciences (75). 46 participants specifically indicated that they have philological education, 11 participants studied or graduated in computer science, and 15 participants selected the option `Other'.

\section{Results and discussion}
\subsection{Correctness of the Answer}
\begin{figure}[!htp]
    \centering
    \ifArxiv\includegraphics[width=\linewidth]{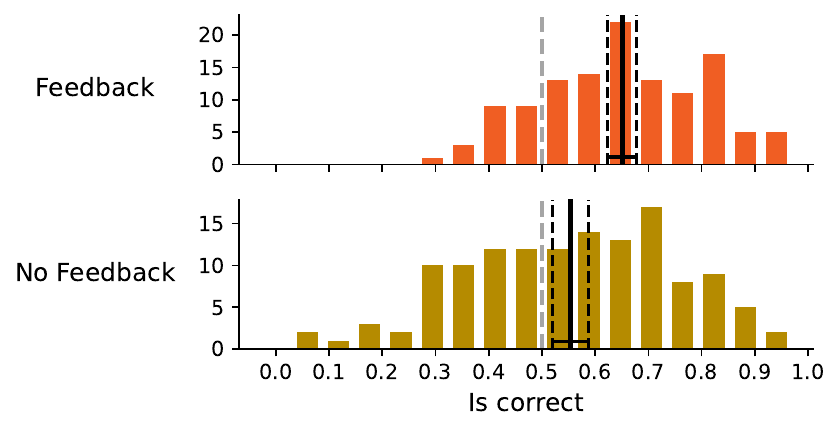}\fi
    \caption{\bf The distribution of the correctness of the answers.}
    \label{fig:correctness_overall}
\end{figure}

Firstly, we present the analysis of the overall correctness rate among the feedback and no-feedback groups using bootstrapped 95\% confidence intervals. Correctness of the answer is a binary variable that indicates whether a participant correctly assigned which text was AI-generated and which was human-written in each trial (1 for correct, 0 for incorrect). The distribution of this variable can be seen in Fig \ref{fig:correctness_overall}.

We observed a substantial and statistically significant difference between feedback and no feedback groups. The no feedback group showed considerably lower performance, with an average correctness rate of 55.4\% (95\% CI: 52.0\%--58.7\%), while the feedback group achieved a higher average success rate of 65.1\% (95\% CI: 62.4\%--67.8\%). These findings confirm hypotheses 1 and 2: participants were able to distinguish AI-generated texts from human-written ones at above-chance levels, though no feedback group performed only slightly above chance.

An interesting phenomenon appears at the left tail of the no-feedback distribution, where several participants were consistently wrong. This likely occurred because these participants adhered to certain strategies that were fundamentally flawed. We do not observe this pattern in the feedback group, presumably because such erroneous strategies were abandoned early in the experiment once participants received corrective feedback.

\subsection{Mixed effects logistic regression}
\label{Model}
Given that the target variable was binary (participants could either respond correctly or incorrectly for each word pair), we employed mixed effects logistic regression. This statistical approach is well-suited for data with non-independent observations --- such as in our study, where each participant provided multiple responses. Mixed effects models allowed us to account for participant-specific variables, including demographic characteristics and individual attitudes toward AI.

For this analysis, the statistical software Jamovi (under \url{https://www.jamovi.org/}, last retrieved 2023/12/29) was utilized, using GAMLj package (\url{https://gamlj.github.io/glmmixed_example1.html}, detailed specification of the models and other settings can be accessed in the Supporting Information files \ref{S2_DataAndScripts}). The analysis focused on the target variable --- correctness of the answer on each trial.

Since we discovered that there was a significant difference in the correctness of the answers between the two groups (feedback and no feedback), we conducted two separate mixed model analyses, one for each group. This decision was based on the expectation that participants in the feedback condition would improve over time, showing a different trajectory in both performance and confidence compared to the no feedback group. Complete results from both models are provided in the tables \ref{tab:learners model} and \ref{tab:nonlearners model}. We summarize the key findings and answer the researched questions raised in \ref{subsec:researchquestion}.

\begin{table*}[ht!]
\begin{adjustwidth}{-2.25in}{0in}
\centering
\caption{
{\bf Feedback group: Mixed effects logistic regression results with estimates, confidence intervals, exponentiated coefficients, and significance levels.}
}
\label{tab:learners model}
\begin{tabular}{|l+l|l|l|l|l|l|l|l|l|}
\hline
&&&\multicolumn{2}{l|}{\bf Conf. Interval} & &\multicolumn{2}{l|}{\bf Conf. Interval} &&\\
{\bf Variable} &{\bf Estimate} &{\bf SE} &{\bf Lower} &{\bf Upper} &{\bf Exp(B)} &{\bf Lower} &{\bf Upper} &{\bf z} &{\bf p }\\  \thickhline

(Intercept)&1.75347&0.3499&1.06769&2.4393&5.775&2.9087&11.464&5.0114&\textbf{\textless0.001}\\\hline
Confidence&0.08892&0.0517&-0.01249&0.1903&1.093&0.9876&1.21&1.7186&0.086\\\hline
More readable is human&1.22118&0.2768&0.67873&1.7636&3.391&1.9714&5.834&4.4123&\textbf{\textless0.001}\\\hline
Trial order&0.05304&0.0132&0.02712&0.079&1.054&1.0275&1.082&4.0111&\textbf{\textless0.001}\\\hline
Reaction time log&0.01424&0.0935&-0.16893&0.1974&1.014&0.8446&1.218&0.1523&0.879\\\hline
Dim. Euclidean distance&-0.00341&0.0581&-0.11732&0.1105&0.997&0.8893&1.117&-0.0586&0.953\\\hline
Dim1 difference&-0.08284&0.1077&-0.29387&0.1282&0.92&0.7454&1.137&-0.7694&0.442\\\hline
Dim2 difference&-0.28056&0.1519&-0.5783&0.0172&0.755&0.5608&1.017&-1.8469&0.065\\\hline
Dim3 difference&0.06608&0.0449&-0.02202&0.1542&1.068&0.9782&1.167&1.4701&0.142\\\hline
Dim4 difference&0.04319&0.0525&-0.05962&0.146&1.044&0.9421&1.157&0.8234&0.410\\\hline
Dim5 difference&-0.03025&0.0632&-0.15419&0.0937&0.97&0.8571&1.098&-0.4784&0.632\\\hline
Dim6 difference&0.07922&0.0393&0.00224&0.1562&1.082&1.0022&1.169&2.017&\textbf{0.044}\\\hline
Dim7 difference&-0.04552&0.0509&-0.14532&0.0543&0.956&0.8647&1.056&-0.894&0.371\\\hline
Dim8 difference&0.01752&0.0417&-0.06414&0.0992&1.018&0.9379&1.104&0.4205&0.674\\\hline
Age&0.00393&0.0108&-0.01723&0.0251&1.004&0.9829&1.025&0.3639&0.716\\\hline
Education&-0.07256&0.1334&-0.33404&0.1889&0.93&0.716&1.208&-0.5439&0.586\\\hline
Gender male - female&-0.08266&0.2068&-0.48794&0.3226&0.921&0.6139&1.381&-0.3998&0.689\\\hline
Reading disorder&0.72202&0.5513&-0.35851&1.8025&2.059&0.6987&6.065&1.3097&0.190\\\hline
Study field: human - cs&-1.70755&0.8003&-3.27602&-0.1391&0.181&0.0378&0.87&-2.1338&\textbf{0.033}\\\hline
Study field: nature - cs&-1.68028&0.8012&-3.25057&-0.11&0.186&0.0388&0.896&-2.0972&\textbf{0.036}\\\hline
Study field: other - cs&-1.27117&0.885&-3.00583&0.4635&0.281&0.0495&1.59&-1.4363&0.151\\\hline
Study field: philology - cs&-1.17388&0.7908&-2.72376&0.376&0.309&0.0656&1.456&-1.4845&0.138\\\hline
Frequency of AI usage&0.05217&0.0625&-0.0704&0.1747&1.054&0.932&1.191&0.8342&0.404\\\hline
AI attitudes: improvement&0.09082&0.0725&-0.05124&0.2329&1.095&0.9501&1.262&1.253&0.210\\\hline
AI attitudes: risk&0.01045&0.0598&-0.10683&0.1277&1.011&0.8987&1.136&0.1746&0.861\\\hline
AI attitudes: quality&0.09781&0.0672&-0.03396&0.2296&1.103&0.9666&1.258&1.4548&0.146\\\hline
AI attitudes: creativity&0.04339&0.0644&-0.0829&0.1697&1.044&0.9204&1.185&0.6734&0.501\\\hline
AI attitudes: understanding&0.06126&0.0569&-0.05023&0.1728&1.063&0.951&1.189&1.0769&0.282\\\hline
AI attitudes: grammar&-0.01654&0.063&-0.14001&0.1069&0.984&0.8693&1.113&-0.2625&0.793\\\hline
AI attitudes: avoidance&0.12247&0.0533&0.01793&0.227&1.13&1.0181&1.255&2.2961&\textbf{0.022}\\\hline
AI attitudes: dumbing&0.01103&0.054&-0.09478&0.1168&1.011&0.9096&1.124&0.2042&0.838\\\hline
AI attitudes: fascination&0.01445&0.0739&-0.13032&0.1592&1.015&0.8778&1.173&0.1956&0.845\\\hline
\end{tabular}
\begin{flushleft}
Table notes: Estimates and exponentiated coefficients (Exp(B)) are reported with 95\% confidence intervals. Bold p-values indicate statistical significance (p \textless 0.05).
\end{flushleft}
\end{adjustwidth}
\end{table*}

\begin{table*}[ht!]
\begin{adjustwidth}{-2.25in}{0in}
\centering
\caption{
{\bf No-feedback group: Mixed effects logistic regression results with estimates, confidence intervals, exponentiated coefficients, and significance levels.}
}
\label{tab:nonlearners model}
\begin{tabular}{|l+l|l|l|l|l|l|l|l|l|}
\hline
&&&\multicolumn{2}{l|}{\bf Conf. Interval} & &\multicolumn{2}{l|}{\bf Conf. Interval} &&\\
{\bf Variable} &{\bf Estimate} &{\bf SE} &{\bf Lower} &{\bf Upper} &{\bf Exp(B)} &{\bf Lower} &{\bf Upper} &{\bf z} &{\bf p }\\  \thickhline

(Intercept)&1.07430&0.3503&0.38767&1.76092&2.928&1.474&5.818&3.0666&\textbf{0.002}\\\hline
Confidence&-0.13657&0.0482&-0.23113&-0.04202&0.872&0.794&0.959&-2.8310&\textbf{0.005}\\\hline
More readable is human&2.06307&0.2178&1.63617&2.48996&7.870&5.135&12.061&9.4719&\textbf{\textless.001}\\\hline
Trial order&-0.01233&0.0113&-0.03447&0.00981&0.988&0.966&1.010&-1.0917&0.275\\\hline
Reaction time log&-0.08757&0.0832&-0.25057&0.07544&0.916&0.778&1.078&-1.0529&0.292\\\hline
Dim. Euclidean distance&0.09811&0.0504&-7.66e-4&0.19699&1.103&0.999&1.218&1.9448&0.052\\\hline

Dim1 difference&-0.30027&0.0850&-0.46679&-0.13376&0.741&0.627&0.875&-3.5344&\textbf{\textless.001}\\\hline
Dim2 difference&-0.33332&0.1332&-0.59441&-0.07224&0.717&0.552&0.930&-2.5022&\textbf{0.012}\\\hline
Dim3 difference&0.09330&0.0415&0.01203&0.17458&1.098&1.012&1.191&2.2500&\textbf{0.024}\\\hline
Dim4 difference&0.11974&0.0500&0.02172&0.21776&1.127&1.022&1.243&2.3942&\textbf{0.017}\\\hline
Dim5 difference&-0.08746&0.0543&-0.19392&0.01901&0.916&0.824&1.019&-1.6100&0.107\\\hline
Dim6 difference&0.00881&0.0346&-0.05902&0.07664&1.009&0.943&1.080&0.2545&0.799\\\hline
Dim7 difference&0.03550&0.0454&-0.05350&0.12450&1.036&0.948&1.133&0.7817&0.434\\\hline
Dim8 difference&-0.04195&0.0344&-0.10945&0.02555&0.959&0.896&1.026&-1.2181&0.223\\\hline
Age&0.01868&0.0139&-0.00859&0.04594&1.019&0.991&1.047&1.3428&0.179\\\hline
Education&0.08890&0.1292&-0.16431&0.34210&1.093&0.848&1.408&0.6881&0.491\\\hline
Gender male - female&-0.01249&0.1846&-0.37436&0.34938&0.988&0.688&1.418&-0.0676&0.946\\\hline
Gender other - female&-0.4545&0.5167&-1.4673&0.5583&0.635&0.231&1.748&-0.88&0.379\\\hline
Reading disorder&0.71990&0.6945&-0.64126&2.08105&2.054&0.527&8.013&1.0366&0.300\\\hline
Study field: human - cs&-0.38585&0.3722&-1.11542&0.34372&0.680&0.328&1.410&-1.0366&0.300\\\hline
Study field: nature - cs&-0.29247&0.3946&-1.06583&0.48089&0.746&0.344&1.618&-0.7412&0.459\\\hline
Study field: other - cs&-0.36696&0.4746&-1.29717&0.56326&0.693&0.273&1.756&-0.7732&0.439\\\hline
Study field: philology - cs&-0.51098&0.4184&-1.33099&0.30904&0.600&0.264&1.362&-1.2213&0.222\\\hline
Frequency of AI usage&0.01728&0.0623&-0.10473&0.13929&1.017&0.901&1.149&0.2777&0.781\\\hline
AI attitudes: improvement&0.08242&0.0736&-0.06181&0.22666&1.086&0.940&1.254&1.1200&0.263\\\hline
AI attitudes: risk&-0.02545&0.0580&-0.13910&0.08820&0.975&0.870&1.092&-0.4389&0.661\\\hline
AI attitudes: quality&-0.01152&0.0670&-0.14286&0.11981&0.989&0.867&1.127&-0.1720&0.863\\\hline
AI attitudes: creativity&-0.01039&0.0654&-0.13853&0.11774&0.990&0.871&1.125&-0.1590&0.874\\\hline
AI attitudes: understanding&0.07057&0.0631&-0.05313&0.19427&1.073&0.948&1.214&1.1182&0.263\\\hline
AI attitudes: grammar&0.23112&0.0645&0.10469&0.35754&1.260&1.110&1.430&3.5830&\textbf{\textless.001}\\\hline
AI attitudes: avoidance&-0.07768&0.0581&-0.19155&0.03619&0.925&0.826&1.037&-1.3370&0.181\\\hline
AI attitudes: dumbing&0.05286&0.0524&-0.04988&0.15559&1.054&0.951&1.168&1.0084&0.313\\\hline
AI attitudes: fascination&-0.01330&0.0816&-0.17320&0.14661&0.987&0.841&1.158&-0.1630&0.871\\\hline
\end{tabular}
\begin{flushleft}
Table notes: Estimates and exponentiated coefficients (Exp(B)) are reported with 95\% confidence intervals. Bold p-values indicate statistical significance (p \textless 0.05).
\end{flushleft}
\end{adjustwidth}
\end{table*}

\subsubsection{Trial order}

This variable is important for our first and second research question. If we are asking whether the group that received feedback differs from the one that did not, we assume that those with feedback were able to learn. Their abilities should therefore improve over time. The model confirmed that they do. The variable Trial order reached level of significance p \textless\ 0.001 in feedback group (CI 0.027--0.079). The log odds ratio estimate was at 0.053, that means, the later the trial, the higher probability that the participant answers correctly. In the no feedback group, on the other hand, the trial order did not reach the level of significance (p=0.25, estimate -0.012).

\begin{figure}[!htp]
    \centering
    \ifArxiv\includegraphics[width=\linewidth]{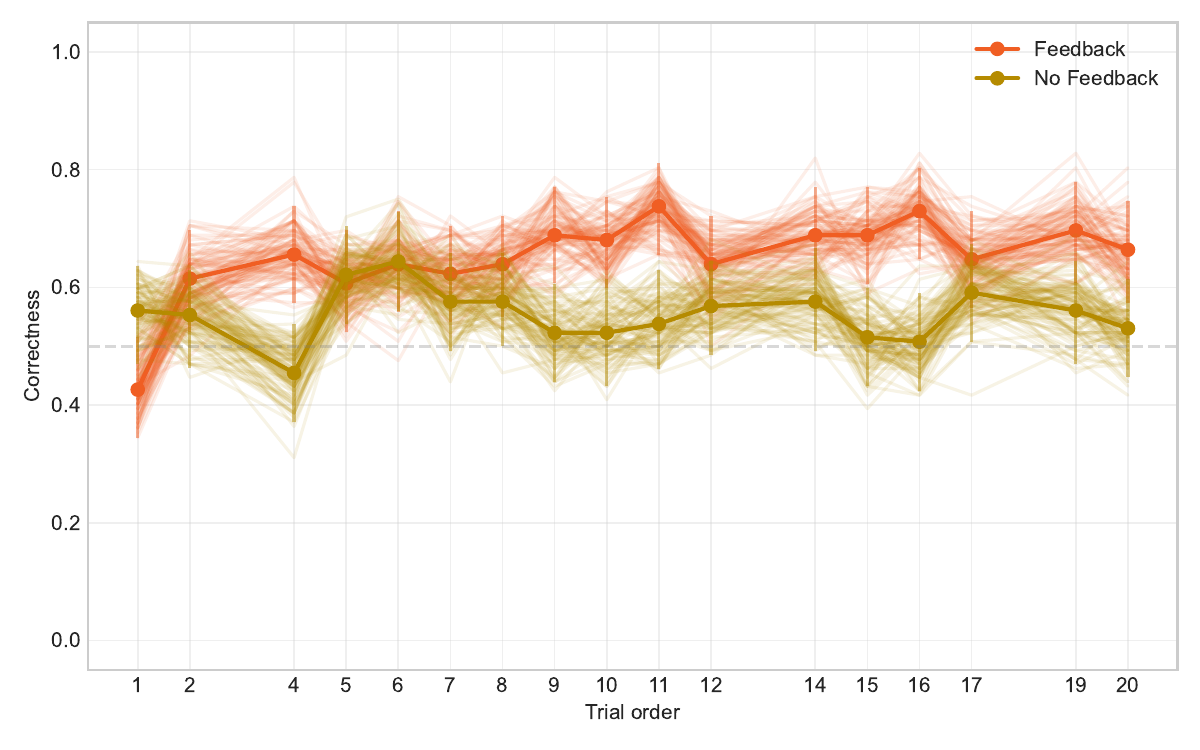}\fi
    \caption{\bf The dependency of correctness on trial order.}
    \label{fig:correctness_trialorder}
\end{figure}

FIG \ref{fig:correctness_trialorder} shows that the correctness rises rather chaotically, which is due to the high variability of the data (half of 254 binary values per datapoint). As the confidence intervals suggest, the initial steep increase in correctness after the first trial for the feedback group might be just noise; however, in the second half of the experiment, the feedback group consistently outperforms the no-feedback group.

\subsubsection{Confidence}
For each pair of texts, participants had to express how confident they were in their answer on a scale of 1--7, i.e., how certain they were that they correctly identified which text was human-written and which was AI-generated.

\begin{figure}[!htp]
    \centering
    \ifArxiv\includegraphics[width=\linewidth]{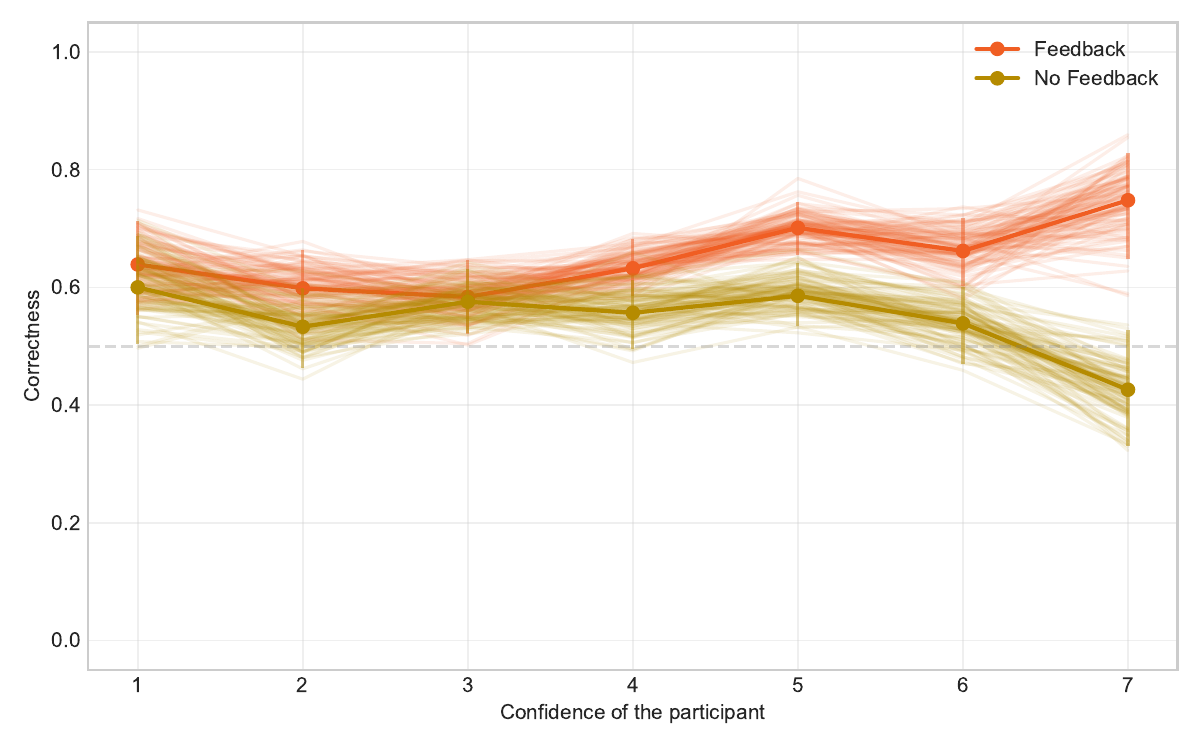}\fi
    \caption{\bf The dependency of correctness on confidence.}
    \label{fig:confidence_correctness}
\end{figure}

There is no significant difference in the average confidence between feedback and no-feedback groups (feedback group: 4.2, CI: 4.0--4.4, no-feedback: 4.3, CI: 4.2--4.5), but the dependence of correctness on confidence is different (see Fig \ref{fig:confidence_correctness}). At low confidence levels, there was little difference between the groups. However, from approximately level five and above, where the feedback group participants were confident, they indeed showed higher correctness. The greatest difference between the groups appears at confidence level seven, where no feedback group participants were overconfident and made the most errors precisely when they were most certain (even below the chance level of 0.5), while those who could update their assumptions via feedback were justifiably confident, achieving decisively better results.

This difference is also reflected in the model: there is no significant effect in the feedback group, but the more no-feedback group participants were confident in their answer, the worse was their correctness rate (p=0.005; estimate -0.14; CI -0.23 -- -0.042).

\begin{figure}[!htp]
    \centering
    \ifArxiv\includegraphics[width=\linewidth]{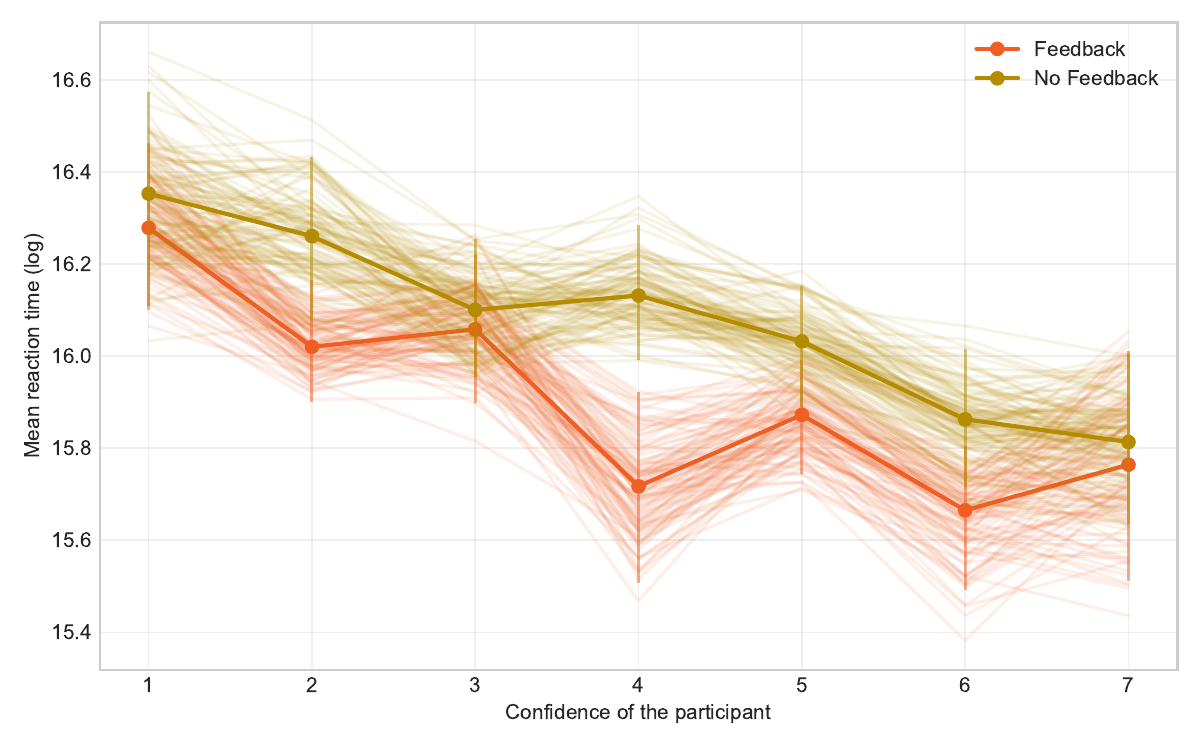}\fi
    \caption{\bf The dependency of reaction time on confidence.}
    \label{fig:confidence_reaction time}
\end{figure}

Furthermore, we examined the relationship of confidence with reaction time and discovered that in situations where participants were more confident in their assessment, the trials were solved quicker, regardless of the group (Fig \ref{fig:confidence_reaction time}).

\begin{figure}[!htp]
    \centering
    \ifArxiv\includegraphics[width=\linewidth]{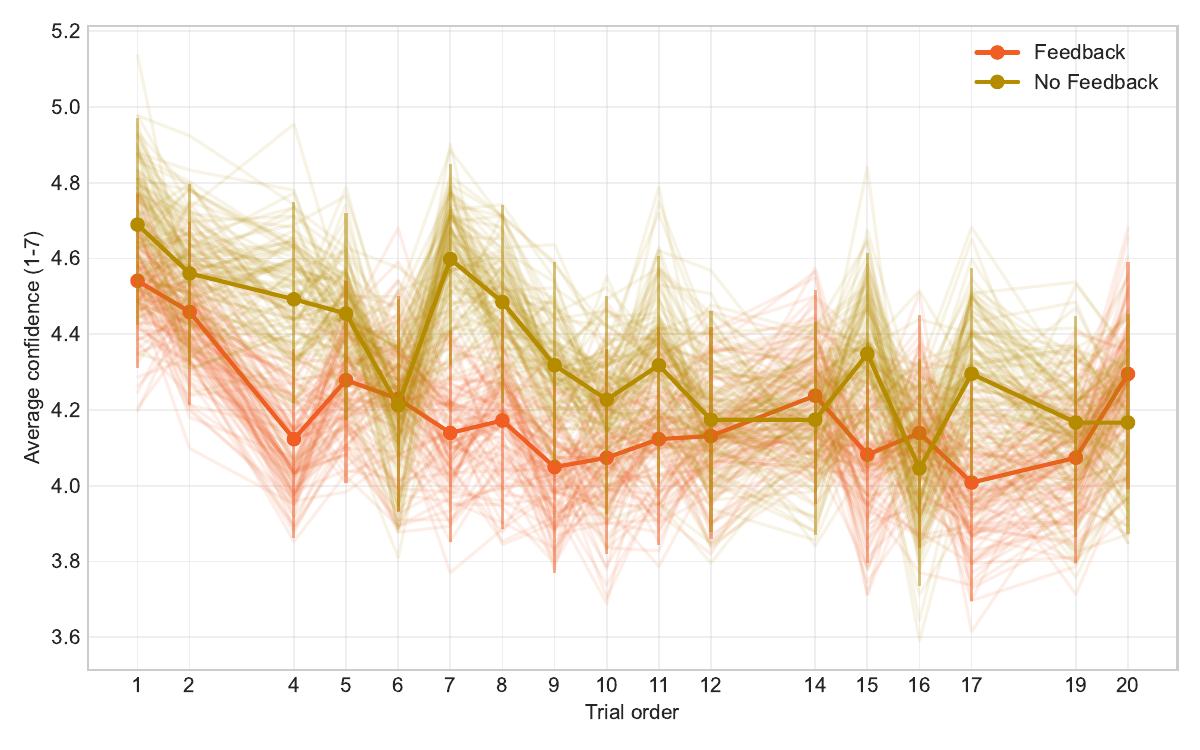}\fi
    \caption{\bf The dependency of confidence on trial order.}
    \label{fig:confidence_trialorder}
\end{figure}

Additionally, we examined the dependence of `confidence' on the trial order, i.e., how the average confidence changes during the experiment session (Fig \ref{fig:confidence_trialorder}). We were interested if we find some kind of development, particularly in the feedback group, but the chart appears rather chaotic.

\subsubsection{Influence of register}
%přejmenovat GLS difference na dimension differences, nebo D1,D2....

Our fourth research question asked if the ability to determine whether a text is written by AI or a human is influence by the genre of the text. To address this, we focused on the `dimension' variables.

Firstly, we look at all the dimensions hollistically, throught Euclidean distance. We utilized the fact that both a human-generated text chunk and an AI-generated text chunk are continuations of the same text, so they should be comparable. It should therefore hold that texts that are stylistically more similar should be harder to distinguish, as this indicates that the AI-generated text chunk is a more credible continuation of the given text. This hypothesis was supported neither for the no feedback group (p = 0.052; estimate: 0.098; CI: 0.000--0.20), neither for the feedback group (p = 0.95; estimate: -0.0034; CI: -0.12--0.11).

Next, we examined the dimensions individually. Although all eight dimensions were included in the analysis, the first two dimensions accounted for the largest proportion of variation in the corpus and were therefore of primary interest. In the tables \ref{tab:nonlearners model} and \ref{tab:learners model}), the variable `Dim difference' indicates how different the two texts from the pair (one human produced, one AI generated) are from each other on a particular dimension. In other words, how much the AI-text `shifted' from the original human text on the particular dimension. A positive estimate indicates a shift toward the positive pole of the dimension. For example, in `Dim1 difference', a positive estimate would suggest a shift from the `static' pole toward the `dynamic' pole. The associated p-value indicates whether this shift significantly affected the probability of correctly identifying the AI-generated text.

Interestingly, we found a stark difference between the feedback and no feedback groups. While the shift in dimensions did not have much of an effect on the feedback group, it had a significant effect on the no feedback group. Among the feedback group, only one stylistic dimension showed a significant effect: the sixth dimension, which captures the tendency of a text to focus on general qualities versus specific referents. For Dim6 difference, the effect was significant, althought the p-value was just above the significance level (p = 0.044; estimate: 0.079; CI: 0.0022--0.16). That indicates that when the AI-generated continuation shifted the original text from a more particular style toward a more general one, the participants who got feedback were more likely to correctly identify it as AI-generated.

On the other hand, in no feedback group, differences in the first four dimensions of variation reached the level of significance. For Dim1 difference, the effect was highly significant (p \textless\ 0.001; estimate: -0.30; CI: -0.47 -- -0.13). That indicates that when the AI-generated text was more static than the original human-written continuation, participants were more likely to identify it correctly. One possible explanation is that, without feedback, participants may rely on prior assumptions about AI writing --- namely, that AI tends to produce static texts characterized by a high density of nominal or phrasal constructions. When such texts are presented alongside more dynamic human continuations (rich in verbal constructions), participants may be more inclined to attribute the static text to AI.

A similar effect was found for Dim2 difference (p = 0.012; estimate: -0.33; CI: -0.59 – -0.072). That means, when the AI-generated continuation shifted from the spontaneous pole toward the prepared pole, it was more likely to be correctly identified. This finding supports the idea that the no feedback group associate prepared, formal styles --- such as those typical of scientific or administrative texts --- with AI. When paired with a more spontaneous and subjective human continuation (e.g., featuring dialogue or narrative elements), the contrast likely reinforces the attribution of the more prepared text to AI.

Significant effects were also found in no feedback group for the third and fourth dimensions. When the AI-generated continuation showed a shift from lower to higher cohesion, participants were more likely to correctly identify it as AI-generated (p = 0.024; estimate: 0.093; CI: 0.012--0.17). Similarly, a shift from monothematic to polythematic content was associated with higher accuracy in AI identification (p = 0.017; estimate: 0.12; CI: 0.022--0.22).

These findings suggest that the participants lacking feedback relied on general assumptions about AI-generated texts --- namely, that they are static, prepared, highly cohesive, and polythematic. When a text matched these expectations and was contrasted with a more dynamic, spontaneous, or thematically focused text, participants tended to attribute authorship to AI.

Those participants who received immediate feedback throughout the task could refine these assumptions. As a result, they no longer treated such stereotypical stylistic features as definitive cues for identifying AI-generated texts.

\subsubsection{Percieved readability}
For each pair of texts, participants were asked which text they found easier to read. The variable `readability' in the model indicates whether the participant thought the human-written text was the more readable one of the pair. 

\begin{figure}[!htp]
    \centering
    \ifArxiv\includegraphics[width=\linewidth]{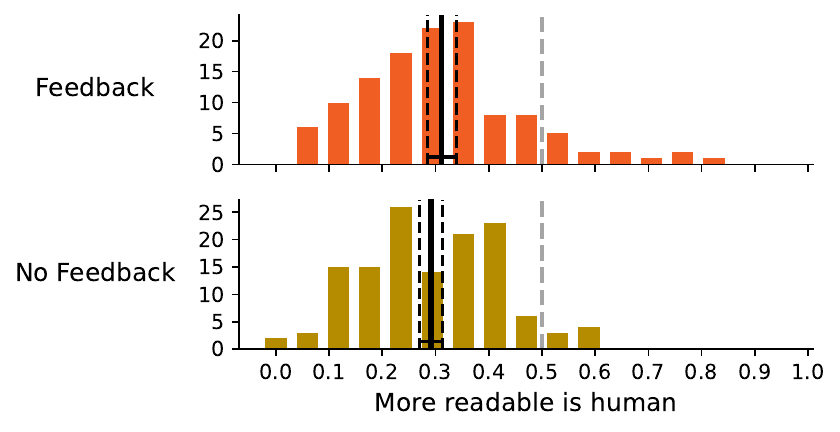}\fi
    \caption{{\bf The distribution of the readability.} The charts shows the frequency at which a human written text is considered more readable.}
    \label{fig:readability_overall}
\end{figure}

As can be seen in Fig \ref{fig:readability_overall}, participants systematically considered AI-written texts more readable (only approximately 30\% of human-produced texts were considered more readable than AI-generated ones), with no significant difference between the feedback / no feedback groups.

\begin{figure}[!htp]
    \centering
    \ifArxiv\includegraphics[width=\linewidth]{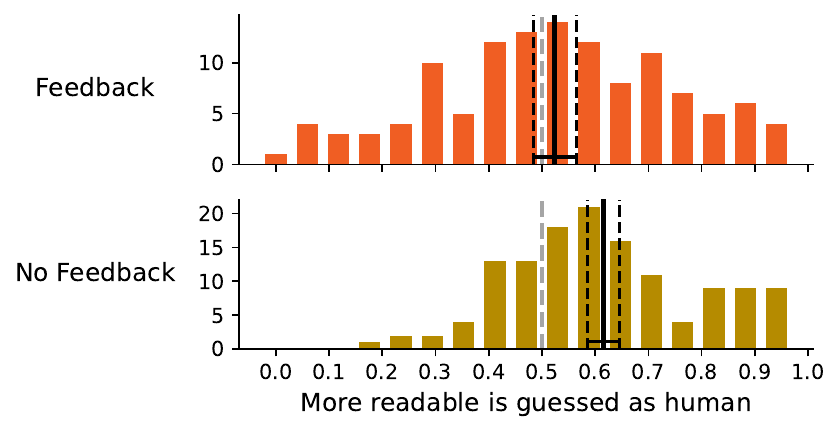}\fi
    \caption{{\bf The distribution of the readability.} The charts shows the frequency at which the text that is considered more readable is also considered to be written by a human.}
    \label{fig:readability_guessed_overall}
\end{figure}

\begin{figure}[!htp]
    \centering
    \ifArxiv\includegraphics[width=\linewidth]{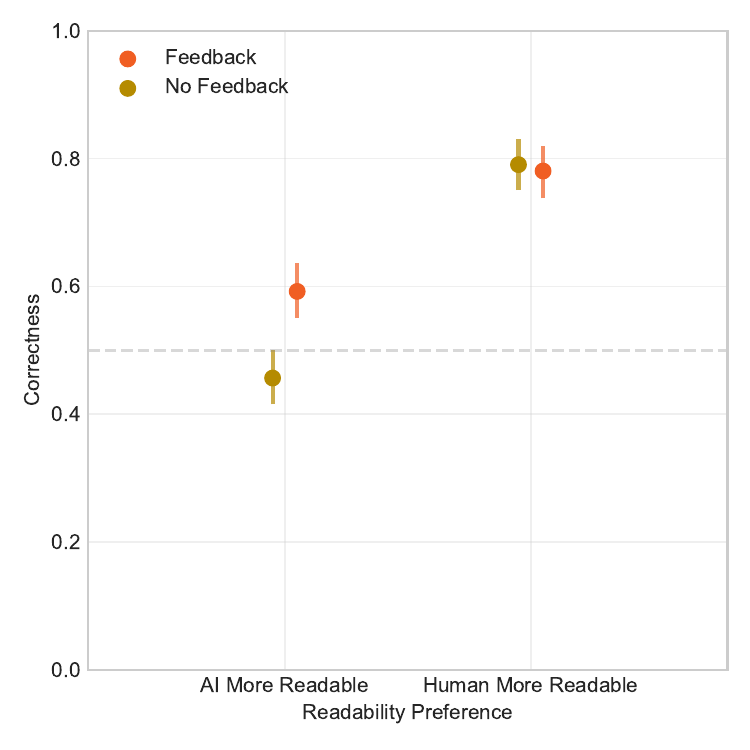}\fi
    \caption{{\bf Correctness by Feedback and Readability.} When participants considered the human written text to be more readable, they were more likely to assing them correctly as human written.}
    \label{fig:readability_correct}
\end{figure}

Participants assign AI generated texts greater readability but at the same time suppose that the more readable texts are written by human (Fig \ref{fig:readability_guessed_overall}). This tendency is especially pronounced in the no feedback group, the feedback group managed to learn to some extent that this assumption is false and that it leads to incorrect answers. The feedback effect can be seen in Fig \ref{fig:readability_correct}, which shows that when participants think that the human written text is more readable, they are more likely to assing them correctly
as human written, but when they think that the AI generated text is more readable, they are less likely to assign them correctly as AI generated, and that this tendency is more strong for no feedback group. 

This finding is also visible in the mixed model: the variable \emph{More readable is human} has significant results for both groups, but the effect size is larger for the no feedback group (feedback: p \textless\ 0.001; estimate 1.22; CI 0.68--1.76; no feedback: p \textless\ 0.001; estimate 2.06; CI  1.64--2.49).

\subsubsection{Attitudes towards AI}

In the survey before the experiment, we asked participants to answer questions about their attitudes and believes towards AI. We hypothesized that those with positive attitudes will achieve better results. However, we found no systematic correlation between the attitudes and the ability to correctly asign AI texts.

In the feedback group, only one statement reached the level of significance: ``I prefer to avoid using AI when writing important texts" (p=0.022; estimate 0.12; CI 0.018--0.23). People who agreed with this statement tended to have better correctness rate than people who dissagreed. In the no feedback group, the only statement reaching level of significance was: ``AI produces grammatically correct texts." When people agreed with this statement, they tended to have a better correctness rate than when disagreed (p \textless\ 0.001; estimate 0.23; CI 0.10--0.36). This finding is intuitively interpretable: if a participant believes that AI produces grammatically incorrect texts, they may struggle more in this task, as AI-generated texts in our experiment were, in fact, largely grammatically correct.

However, since no consistent pattern emerged across attitude statements and groups, these findings should be interpreted with caution. Rather than drawing firm conclusions, we suggest that these effects point to potentially directions for future research. In particular, investigating how beliefs about grammar and the formal writing capabilities of AI influence performance could be fruitful. For now, we conclude that our fourth hypothesis --- ``Individuals with a more positive attitude toward AI will be more successful in determining authorship." --- was not supported by the data.

\subsubsection{Frequency of usage of AI}

We did not find a significant effect of AI usage frequency in either group. Before the experiment, participants answered the question ``How often do you use large language models?", which we treated as an ordinal variable. However, this measure was not a significant predictor of success in the AI text recognition task. In other words, participants who reported frequent interactions with AI did not perform better than those with less experience.

Therefore, our fifth hypothesis --- ``Individuals who interact with AI more frequently will be more successful in determining authorship." --- was not supported by the data.

In the following sections, we present additional findings revealed by the mixed model analysis.

\subsubsection{Demographic details}
Firstly, the model uncovered that gender, age, and education level did not have a significant effect on performance in either group. We also examined participants' field of study and found no consistent effect on correctness. In the learners group, participants with a background in computer science performed better than those from the humanities or natural sciences. However, it is important to note that fewer than 11 participants in total reported computer science as their field of study, so this finding should be interpreted with caution.

%\begin{figure}[!htp]
%    \centering
%    \includegraphics[width=\linewidth]{Supplementary_information/Demographics/correctness_Demographics-field_merged.pdf}
%    \caption{The distribution of the correctness of the answers depending on the field of study of the participants (cs = computer science; nature = nature science ; human = humanities; philology = philological fields; other = other fields.}
%    \label{fig:fieldofstudy_correctness}
%\end{figure}

\section*{Conclusion}

The key takeaway from this study is that although everyday interactions with AI do not inherently enhance individuals' abilities to differentiate between human and AI-generated texts (as evidenced by the negligible effect of participants' previous AI interaction time), this skill can be effectively learned through targeted training with explicit feedback loops.

Participants initially hold numerous assumptions about the stylistic features of AI-generated texts. They also incorrectly anticipate that more readable texts are typically human-authored. Feedback significantly aids in mitigating these misconceptions.

%These findings suggest that the participants lacking feedback relied on general assumptions about AI-generated texts --- namely, that they are static, prepared, highly cohesive, and polythematic.
Specifically, participants assume that AI will produce texts that resemble static, cohesive, and prepared genres (such as administrative or scientific texts) rather than texts resembling spontaneous, dynamic genres (such as dialogues or narrative prose). While these assumptions occasionally help when an AI-generated text aligns with them, they frequently lead to errors when human texts deviate from expected human characteristics. Feedback enables participants to appropriately calibrate the extent to which they should rely on these preconceived notions. Similarly, feedback helps participants correct their erroneous belief that human-generated texts are inherently more readable.

Feedback also facilitates more accurate self-assessment of participants' abilities to discriminate between human and AI-generated texts. The overall confidence levels remain consistent across both groups; however, participants receiving feedback exhibit more accurate confidence calibration, being confident primarily when their judgments are indeed correct. Notably, individuals without feedback make the most errors precisely when they feel most confident about their judgments. 

Our findings have practical implications as well. Specifically, our experimental software can be easily adapted into a practical tool for educators, enabling them to assess and enhance their ability to recognize AI-generated content.

%\nameref{S1_Appendix}.

\section*{Supporting information}

% Include only the SI item label in the paragraph heading. Use the \nameref{label} command to cite SI items in the text.
\paragraph*{S1 File.}
\label{S1_Protocol}
{\bf Protocol.} Detailed description of the experiment and all methods and techniques to obtain and analyze the data, descriptive statistics of the data. \ifArxiv Available at \url{https://osf.io/uhzj9/}.\fi

\paragraph*{S2 File.}
\label{S2_DataAndScripts}
{\bf Data and scripts.} All data and scripts used in the study, some additional descriptive statistics, scalable formate figures.\ifArxiv Also available at \url{https://osf.io/uhzj9/}.\fi

\ifArxiv\section*{Funding}
Jiří Milička was supported by Czech Science Foundation Grant No. 24-11725S, gacr.cz (\emph{Large language models through the prism of corpus linguistics}). 

This work was supported by the project \emph{Human-centred AI for a Sustainable and Adaptive Society} (reg. no.: CZ.02.01.01/00/23\_025/0008691), co-funded by the European Union (specifically the development of the experimental software). 

Anna Marklová was supported by Charles University Grant PRIMUS/25/SSH/010 (\emph{Sensitivity to register variability: A combination of corpus and experimental methodology}).\fi

\section*{Acknowledgments}
We thank Jan Chromý for providing access to the participant pool within the LABELS laboratory and for assistance with the administration of the experiment.

\nolinenumbers

% Either type in your references using
% \begin{thebibliography}{}
% Text
% \end{thebibliography}
%
% or
%
% Compile your BiBTeX database using our plos2015.bst
% style file and paste the contents of your .bbl file
% here. See http://journals.plos.org/plosone/s/latex for 
\bibliography{reference}

\end{document}